\documentclass[10pt,twocolumn,letterpaper]{article}
\usepackage{cvpr}
\usepackage{epsfig}
\usepackage{graphicx, calc}
\usepackage{amsmath}
\usepackage{amssymb}
\usepackage{subcaption}
\usepackage{booktabs}
\usepackage{array}
\usepackage{tabularx}
\usepackage{multirow}
\usepackage[numbers,sort,compress]{natbib} % sort citations
\usepackage{enumitem}
\usepackage[table,xcdraw]{xcolor}
\usepackage{verbatim}

\usepackage{kotex}
\usepackage[T1]{fontenc}
\usepackage[utf8]{inputenc}
\usepackage{times}
\usepackage{pdfpages}  % to avoid errors

\renewcommand{\paragraph}[1]{\noindent \textbf{#1}}
\makeatletter\renewcommand\paragraph{\@startsection{paragraph}{4}{\z@}
  {.5em \@plus1ex \@minus.2ex}{-.5em}{\normalfont\normalsize\bfseries}}\makeatother

\usepackage[pagebackref=true,breaklinks=true,letterpaper=true,colorlinks,bookmarks=false]{hyperref}

\newcommand{\z}{\phantom{0}}

\newif\ifrone
\ronetrue
\newif\ifsupplementary
\supplementaryfalse
\newif\ifappendix
\appendixtrue
\newcommand{\Table}[1]{Table~\ref{#1}}

\newcommand{\Figure}[1]{Fig.~\ref{#1}}
\newcommand{\Equation}[1]{Eq.~(\ref{#1})}

\newcommand{\Section}[1]{Sec.~\ref{#1}}
\newcommand{\ignore}[1]{}

\newcommand{\basemodel}{Base Model}
\newcommand{\vmuse}{MUVE}
\newcommand{\howto}{HowToWorld}
\newcommand{\howtoshort}{HowToW}
\newcommand{\howtoen}{\howtoshort{}-En}
\newcommand{\howtofr}{\howtoshort{}-Fr}

\newcommand{\howtotext}{\howtoshort{}-Text}

\newcommand{\howtotextall}{\howtoshort{}-Text-\{En,Fr,Ko,Ja\}}

\newcommand{\tablestyle}[2]{\setlength{\tabcolsep}{#1}\renewcommand{\arraystretch}{#2}\centering\footnotesize}

\newlength\myheight
\newlength\mydepth
\settototalheight\myheight{Xygp}
\newcommand*\inlinegraphics[1]{%
	\settototalheight\myheight{Xygp}%
	\settodepth\mydepth{Xygp}%
	\raisebox{-\mydepth}{\includegraphics[height=\myheight]{#1}}%
}

\cvprfinalcopy % *** Uncomment this line for the final submission
\newcommand{\myurl}[1]{\url{#1}}

\def\cvprPaperID{5854}

\ifcvprfinal\fi

\begin{document}

%%%%%%%%% TITLE
\title{Visual Grounding in Video for Unsupervised Word Translation}

\renewcommand{\thefootnote}{\fnsymbol{footnote}}
\author{Gunnar A. Sigurdsson$^3\footnotemark[1]$ \ \ \ \ 
Jean-Baptiste Alayrac$^1$ \ \ \ 
Aida Nematzadeh$^1$ \ \ \ 
Lucas Smaira$^1$  \\
Mateusz Malinowski$^1$ \ \ \
Jo\~ao Carreira$^1$ \ \ \ 
Phil Blunsom$^{1,2}$ \ \ \
Andrew Zisserman$^{1,2}$ \\
\\
$^1$DeepMind \\
$^2$Department of Engineering Science, University of Oxford \\
$^3$Carnegie Mellon University \\
\myurl{github.com/gsig/visual-grounding} %\myurl{url.com/at/the/web}
}

\ifsupplementary  % just supplementary
    \makeatletter
    \let\oldtitle\@title
    \makeatother
    \title{Supplementary Material: \\ \oldtitle }
    \author{}
    \maketitle
    
    \setcounter{section}{6}
    \section{Appendix}
    \noindent This appendix contains the following content:

\begin{enumerate}[itemindent=\parindent,itemsep=0em]
    \item [\ref{sec:results_for_recall}.]\  \nameref{sec:results_for_recall}
    \item [\ref{sec:text_evaluation_sets}.]\  \nameref{sec:text_evaluation_sets}
\end{enumerate}

\subsection{Results for Recall@10}
\label{sec:results_for_recall}

In \Table{tab:vision2} and \Table{tab:vanilla2} we present numbers for Recall@10 to complement the Recall@1 tables in the paper.
Overall the same trend is observed with that metric.

\begin{table}[hb]
    \tablestyle{6pt}{1.05}
	\begin{tabularx}{\linewidth}{p{1em}Xlccccc}
		\toprule
		\multicolumn{2}{l}{English-French} & \multicolumn{2}{c}{Dictionary} && \multicolumn{2}{c}{Simple Words} \\
		\cmidrule{3-4}
		\cmidrule{6-7}
		&& All & Visual && All & Visual \\
		\midrule
		1) & Random Chance & \z0.7 & \z1.6 && \z1.0 & \z2.2 \\
		2) & Video Retrieval & 16.2 & 17.3 && 29.5 & 43.2  \\
		3) & \basemodel{} & 21.3 & 33.7 && 54.0 & 74.0 \\
		4) & \vmuse{} & \textbf{45.7} & \textbf{60.7} && \textbf{79.7} & \textbf{89.5} \\
		\bottomrule
	\end{tabularx}
	\caption{The performance of our models and the baselines measured as Recall@10 on the \emph{Dictionary} and \emph{Simple Words} benchmarks.}
	\label{tab:vision2}
\end{table}

\begin{table}[hb]
    \tablestyle{4pt}{1.05}
	\begin{tabularx}{\linewidth}{p{1em}Xlcccccc}
		\toprule
		\multicolumn{2}{l}{\multirow{2}{*}{Dictionary}} & \multicolumn{2}{c}{En-Fr} && En-Ko && En-Ja \\
		\cmidrule{3-4} \cmidrule{6-6} \cmidrule{8-8}
		& & All & Visual && All && All \\
		\midrule
		1) & Iterative Procrustes                &  \z0.8         &\z0.9          && \z1.1         && \z0.6 \\
		2) & MUSE~\cite{conneau2017word}  &  42.3          & 57.8          && 23.9          && 23.5 \\
		3) & VecMap~\cite{artetxe2018acl}        &  44.1          & 60.7          && 26.8          && 27.3 \\
		4) & \vmuse{}                            &  \textbf{45.7} & \textbf{60.7} && \textbf{33.4} && \textbf{31.2} \\
		\midrule
		5) & Supervised                          &   80.1         & 84.0          && 72.1          && 68.3 \\
		\bottomrule
	\end{tabularx}
	\caption{Performance of our and text-based methods across different language pairs. We report Recall@10 on the \emph{Dictionary} dataset.}
	\label{tab:vanilla2}
\end{table}

\iftrue
    % bundle compiled pdf with international fonts
    \newcommand{\fakesubsection}[2]{%
        \begin{flushright}
            \textcolor{white}{\subsection{#1}\label{#2}}
        \end{flushright}
    }
    \newcommand{\addpdf}[2]{%
        \includepdf[pages={#1},pagecommand={#2}]{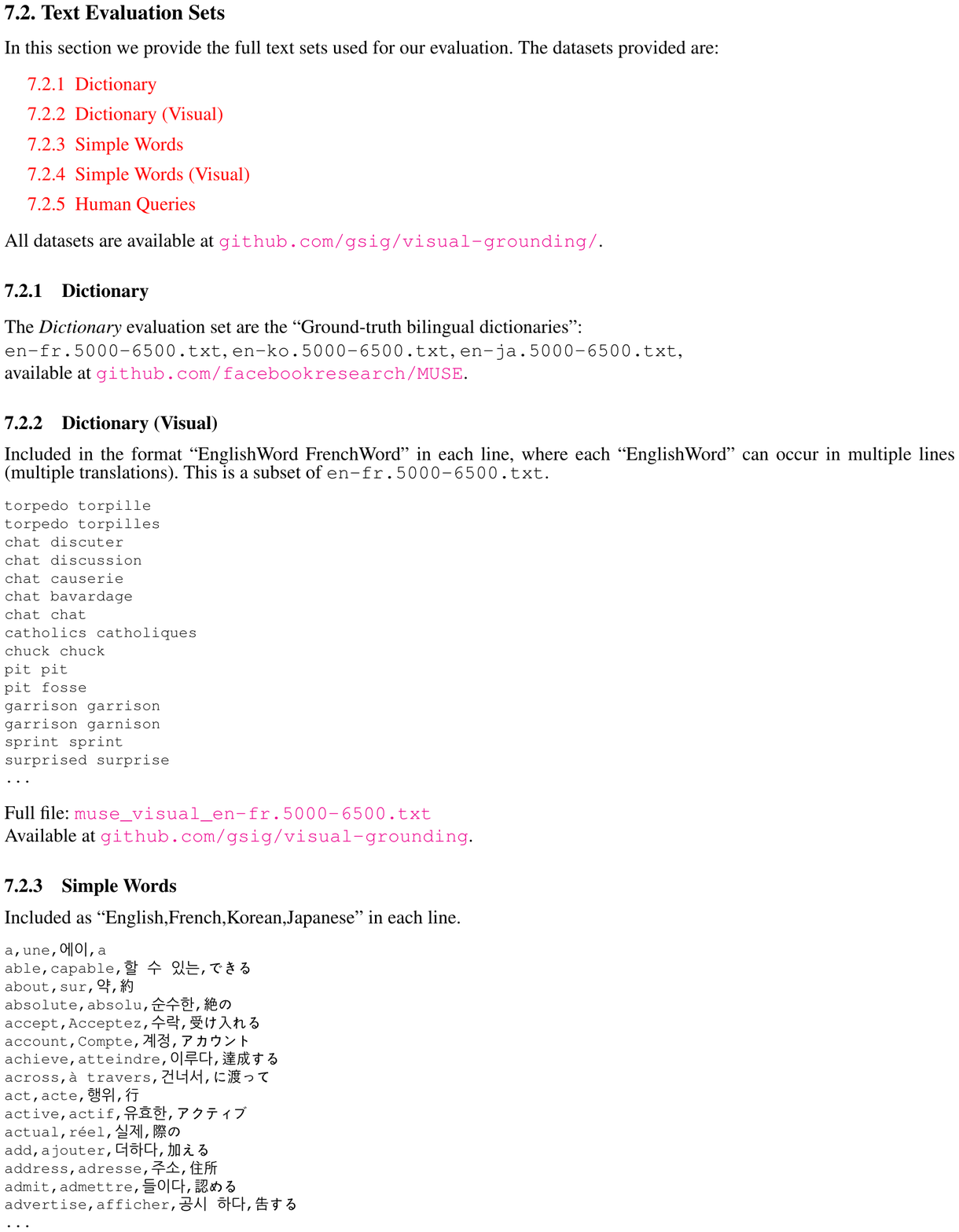}
    }
    \onecolumn
    \addpdf{1}{%
      \fakesubsection{Text Evaluation Sets}{sec:text_evaluation_sets}
      %\fakesubsubsection{Dictionary}{sec:dictionary}
      %\fakesubsubsection{Dictionary (Visual)}{sec:dictionary_visual}
      %\fakesubsubsection{Simple Words}{sec:simple_words}
    }
    \addpdf{2}{%
      %\fakesubsubsection{Simple Words (Visual)}{sec:simple_words_visual}
      %\fakesubsubsection{Human Queries}{sec:human_queries}
    }
\else
    % compile as normal
    % limit verbatim to 15 lines
\makeatletter
\newcounter{VerbatimLineNo}
\def\verbatim@read@file{%
  \stepcounter{VerbatimLineNo}%
  \read\verbatim@in@stream to\next
  \ifnum\value{VerbatimLineNo}>15 
    ...
  \else
    \ifeof\verbatim@in@stream
    \else
      \expandafter\verbatim@addtoline\expandafter{\next}%
      \verbatim@processline
      \verbatim@startline
      \expandafter\verbatim@read@file
    \fi
  \fi
}
\makeatother

\newcommand{\myverbatiminput}[1]{%
  \setcounter{VerbatimLineNo}{0}
  {\footnotesize
    \verbatiminput{datasets/#1}
  }
  \noindent Full file: \myurl{#1}
  \newline Available at \myurl{github.com/gsig/visual-grounding}.
}

\onecolumn
\subsection{Text Evaluation Sets}
\label{sec:text_evaluation_sets}

\noindent In this section we provide the full text sets used for our evaluation. The datasets provided are:
\begin{itemize}[itemindent=\parindent+.5em,itemsep=0em]  % .5em is magic number for subsubsection items
    \item [\ref{sec:dictionary}] \nameref{sec:dictionary}
    \item [\ref{sec:dictionary_visual}] \nameref{sec:dictionary_visual}
    \item [\ref{sec:simple_words}] \nameref{sec:simple_words}
    \item [\ref{sec:simple_words_visual}] \nameref{sec:simple_words_visual}
    \item [\ref{sec:human_queries}] \nameref{sec:human_queries}
\end{itemize}
All datasets are available at \myurl{github.com/gsig/visual-grounding/}.

\subsubsection{Dictionary}
\label{sec:dictionary}

The \emph{Dictionary} evaluation set are the ``Ground-truth bilingual dictionaries'': \newline \texttt{en-fr.5000-6500.txt}, \texttt{en-ko.5000-6500.txt}, \texttt{en-ja.5000-6500.txt},
\newline available at \myurl{github.com/facebookresearch/MUSE}.

\subsubsection{Dictionary (Visual)}
\label{sec:dictionary_visual}
Included in the format ``EnglishWord FrenchWord'' in each line, where each ``EnglishWord'' can occur in multiple lines (multiple translations). This is a subset of \texttt{en-fr.5000-6500.txt}.
\myverbatiminput{muse_visual_en-fr.5000-6500.txt}

\subsubsection{Simple Words}
\label{sec:simple_words}
Included as ``English,French,Korean,Japanese'' in each line.
\myverbatiminput{1000_most_frequent_en_en_fr_ko_ja.csv}

\subsubsection{Simple Words (Visual)}
\label{sec:simple_words_visual}
Included as ``English,French,Korean,Japanese'' in each line.
\myverbatiminput{1000_most_frequent_en_visual_en_fr_ko_ja.csv}

\subsubsection{Human Queries}
\label{sec:human_queries}
Included as ``English,French,Korean,Japanese'' in each line.
\myverbatiminput{howto100m_webqueries_clean_en_fr_ko_ja.csv}
\fi
    \newpage
    {\small
    \bibliographystyle{ieee_fullname}
    \bibliography{cvpr2020gunnar}
    }
\else  % full paper
    \maketitle
    \begin{abstract}
    There are thousands of actively spoken languages on Earth, but a
    single visual world. Grounding in this visual world has the potential
    to bridge the gap between all these languages. Our goal is to use visual grounding to improve unsupervised word mapping between languages. 
    The key idea is to establish a common visual representation between
    two languages by learning embeddings from {\em unpaired} instructional videos narrated in the native language. Given this shared embedding we demonstrate that (i) we can map words between the languages, particularly the `visual' words;  (ii) that the shared embedding provides a good initialization for existing unsupervised text-based word translation techniques, forming the basis for our proposed hybrid visual-text mapping algorithm, MUVE; and (iii) our approach achieves superior performance by addressing the shortcomings of text-based methods -- it is more robust, handles datasets with less commonality, and is applicable to low-resource languages. We apply these methods to translate words from English to French, Korean, and Japanese -- all without any parallel corpora and simply by watching many videos of people speaking while doing things.  
    \end{abstract}

    \footnotetext{\footnotemark[1]Work done while Gunnar was an intern at DeepMind.}
    \section{Introduction}

\begin{figure}[]
\centering
	\includegraphics[width=1.0\linewidth]{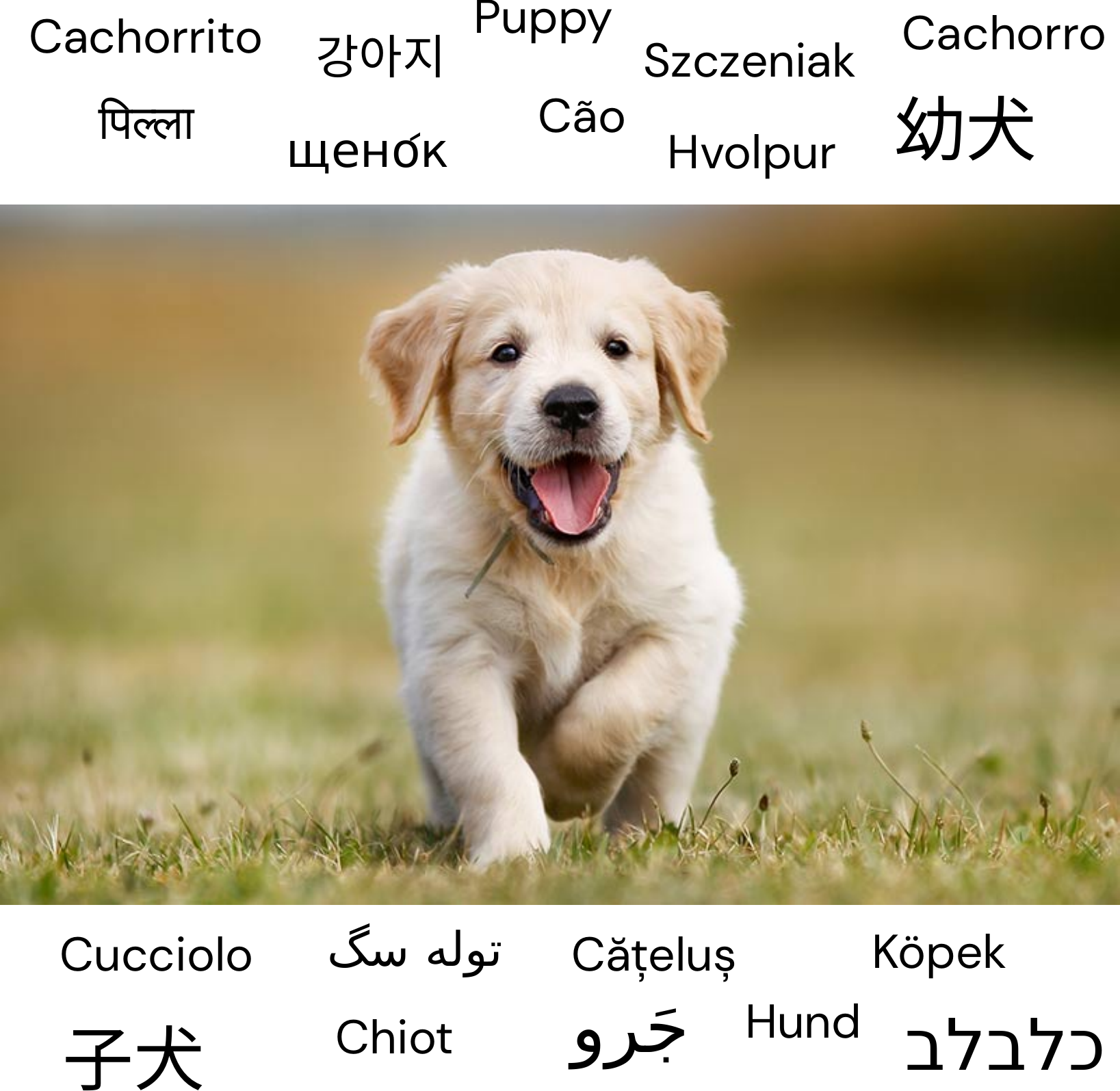}
\caption{Across the world, there are many different ways to refer to \protect\inlinegraphics{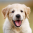}.
	But in the visual domain, a \protect\inlinegraphics{im/small_dog.png} is simply a \protect\inlinegraphics{im/small_dog.png} everywhere on Earth.
	In this work, we leverage this observation to learn to translate words in different languages without \emph{any} paired bilingual data.}
	\label{fig:teaser}
	\vspace{-0.6cm}
\end{figure}

\begin{figure*}[t]
		\centering
		\includegraphics[width=1.0\linewidth]{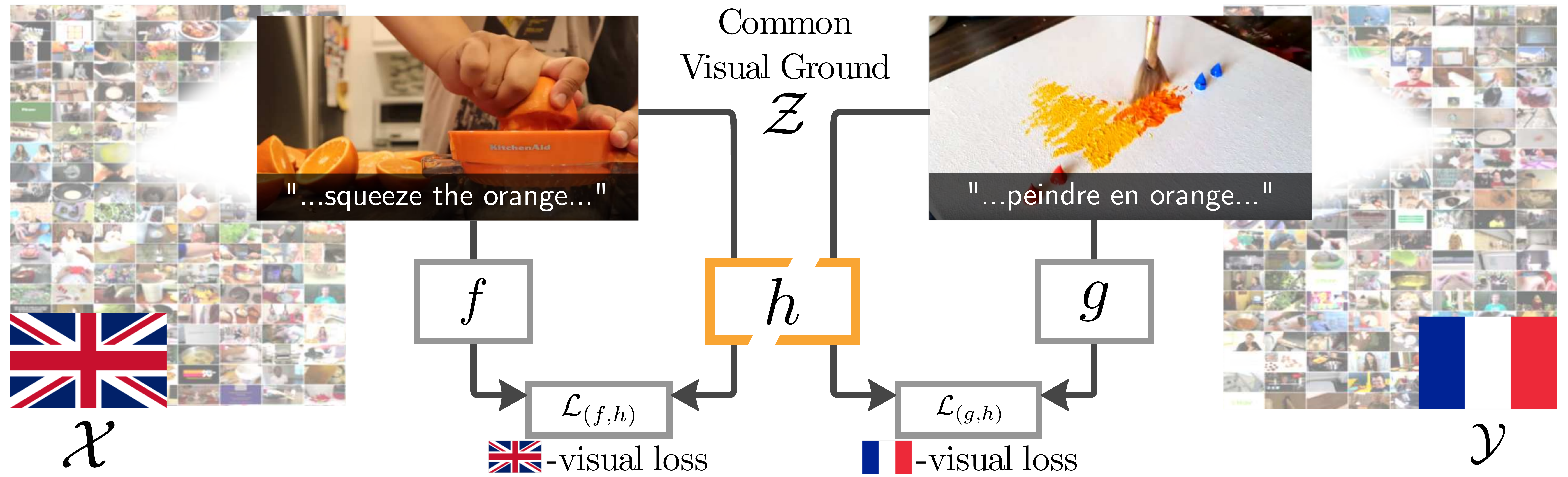}
		\vspace*{-0.5cm}
	\caption{\label{fig:model_full} \small We build on recent advances in video modeling and train an unsupervised system that learns to translate words in multiple languages by grounding the language in video and without any paired data. (``peindre en orange''=``painting in orange''.)
	}
	\vspace{-0.3cm}
\end{figure*}

Children can learn multiple languages by merely observing their environment and interacting with others, without any explicit supervision or instruction; multilingual children do not hear a sentence and its translation simultaneously, and they do not hear a sentence in multiple languages while observing the same situation~\cite{genesee2004dual}.
Instead, they can leverage visual similarity across situations: what they observe while hearing ``the dog is eating'' on Monday is similar to what they see as they hear ``le chien mange'' on Friday. 

We take a first step towards building an unsupervised multimodal translation system by relating the machine translation task to the way children learn multiple languages: we expose the system to videos of people from different countries performing a task while explaining what they are doing in their native languages. 
There are many such videos in YouTube: for example, we can learn how to squeeze orange juice by watching Korean or English videos. Instructional videos tend to look visually similar and the underlying concepts being spoken are often the same. We obtained a large number of such videos and the corresponding subtitles using automatic speech recognition, extending the recent procedure of~\cite{miech19howto100m} to multiple languages.

Working with this data introduces various challenges. First of all, despite significant recent progress, visual understanding in videos is far from solved -- even with the state-of-the-art models, clustering similar activities is not easy. 
Additionally, and in contrast to manually-captioned datasets where words tend to describe the scene, in instructional videos the words correspond to what the instructors are saying. While performing a task, the instructors often talk
about random topics (such as subscriber counts and audience interaction) that do not have any visual relevance.

This paper demonstrates that, despite these challenges, a shared visual representation can facilitate the mapping of different languages at the word level. 
As illustrated in \Figure{fig:model_full}, we propose a model that maps two languages through the visual domain (videos). For English and French, the model correctly translates 28.0\% and 45.3\% of common words and visual words, all by only watching videos. For comparison, a retrieval-based baseline (without sharing the visual representation) achieves 12.5\% and 18.6\% for common words and visual words.

Moreover, we show that our model is more robust than the state-of-the-art unsupervised \emph{text-based} word mapping models which exploit co-occurrence statistics~\cite{artetxe2017acl,conneau2017word}, in terms of sensitivity to (a) the degree to which the two languages differ (\eg, English is more similar to French than Korean), (b) the dissimilarity of the training corpora of the two languages (\eg, English and French Wikipedia are highly similar), and (c) the amount of training data.
Finally, we show that the combination approach (with text-based approaches) is reliable for a large variety of tasks. For example, when the training corpora in French and English are dissimilar (instructional videos in French and Wikipedia in English), our method achieves a 32.6\% recall while that of the text-based ones is less than 0.5\%.

\paragraph{Contributions.}
The contributions are threefold.
\textbf{(i)} We propose a method to map languages through the visual domain using \emph{only} unpaired instructional videos,
\textbf{(ii)} we demonstrate that our method is effective at connecting words in different languages through vision in an unsupervised manner,
and finally \textbf{(iii)} we show that our method can serve as a  good initialization for existing word mapping techniques addressing many shortcomings of \emph{text-based} methods.

    \section{Prior Work}

\paragraph{Bilingual child language acquisition.} 
An open question in the field of bilingual language acquisition is to what extent the systems and representations learned for each language are shared. This sharing can happen for different aspects of language such as grammar, morphology, or the conceptual representations \cite{genesee1989early,de2017bilingual}. For example, bilingual children eventually learn that both ``chien'' and ``dog'' refer to the actual animal dog, but whether and when this representation is shared is a matter of debate. We explore whether sharing the conceptual (visual) representation improve the quality of word translation for different languages.

\paragraph{Unsupervised text-based word alignment.} Words often occur in the same context in different languages -- in both English and French, ``dog'', ``catch'', and ``ball'' co-occur together. Previous work has used this insight to align the embedding space of different languages and use the aligned space to translate words from one language to another language ~\cite{mikolov2013exploiting,lazaridou2015hubness}. 
Earlier work used various degrees of supervision through ground-truth dictionaries or heuristics ~\cite{smith2017offline,hoshen2018non,artetxe2017acl}; recently, fully unsupervised approaches achieved a similar performance on word alignment for different language pairs without any supervision~\cite{artetxe2018acl,conneau2017word}.
However, because these methods take advantage of the similarity between both the language pairs and their training corpora, they are not robust when the languages (or their training corpora) are very different~\cite{sogaard2018limitations,artetxe2018aaai}.

\paragraph{Vision and language.}
There is a growing interest in combining methods developed in computer vision and natural language processing to solve more challenging problems at the intersection of these fields~\cite{kazemzadeh2014referitgame,donahue2015long,malinowski2017ask,hu2019language,vinyals2015show,hermann2019learning,anderson2018vision,shen2018lifelong}.
Grounding language is at the core of the interest of these two communities. It also has a long tradition in symbolic artificial intelligence, where ``meaningless symbols cannot be grounded in anything but other meaningless symbols''~\cite{harnad1990symbol}. 
The same problem of assigning meaning to symbols has been a fruitful research direction in computer vision.
Early work explored weak supervision and the correspondence problem between text annotations and image regions~\cite{Barnard02,Duygulu02},
with more modern approaches exploring joint image-text word embeddings~\cite{Frome13},
or building a language conditioned attention map over the images in caption generation, visual question answering and text-based retrieval~\cite{Xu15,rohrbach2017generating,vinyals2015show,ross2018grounding,anne2016deep,de2017guesswhat,malinowski2018learning,zhu2016visual7w,hendricks17localizing}.
Of particular interest, recent work has focused on multimodal and multilingual settings such as producing captions in many languages, visual-guided translations~\cite{barrault2018findings,elliott2016multi30k,su2019unsupervised,wang2019vatex}, or bilingual visual question answering~\cite{gao2015you}. 
However, these use a paired corpora, \ie same video or images are associated with captions in multiple languages~\cite{wang2019vatex}.
Obtaining paired corpora in several languages is expensive, and does not scale.

\paragraph{Instructional videos.}	
In this work, we rely on instructional videos~\cite{alayrac16unsupervised,Sener_2015_ICCV,yu14instructional} since they can be obtained at scale \emph{without any manual annotation}~\cite{miech19howto100m}: they consist of YouTube videos and their associated narrations which is generated using automatic speech recognition (ASR).
We propose to use instructional videos in different languages to show that we can translate words by only watching and listening to people performing various tasks.

    \section{Unsupervised Multilingual Learning}

We describe our approach for unsupervised multilingual word alignment through grounding in the visual domain~$\mathcal{Z}$.
Our method is \emph{unsupervised} in that it learns the correspondences between two languages $\mathcal{X}$ and $\mathcal{Y}$ (\eg English and French) without \emph{any} parallel (paired) corpora. Instead, we are given two distinct collections of instructional videos, \ie $n$ videos narrated with language $\mathcal{X}$ and another $m$ \emph{different} videos with language $\mathcal{Y}$.
Equipped with this, our goal is to learn to map languages $\mathcal{X}$ and $\mathcal{Y}$ by leveraging the shared visual modality $\mathcal{Z}$ -- the videos.
We evaluate this ability in terms of the accuracy of word translation, \ie how well the vocabulary in one language can be mapped to the other one.

Mapping languages through instructional videos is challenging:
first, learning video-text embeddings from instructional videos is difficult as the speech in these videos is only \emph{loosely} related to the scene.\footnote{\eg only 50\% of captions and videos in HowTo100M are related~\cite{miech19howto100m}.}
Second, in multilingual setting, such errors compound since both languages have this low video-text relevance; moreover, visually similar videos may not be semantically similar.

This challenge \emph{cannot} be addressed by using the similarity of videos to construct a parallel text corpora (see \Figure{fig:noisy}). Instead, following Miech \etal~\cite{miech19howto100m}, we learn a joint (monolingual) video-text embedding space from instructional videos. 
We extend the training strategy to the multilingual case by defining the following objective:
\begin{equation}
\min_{f,g,h} \underbrace{\mathcal{L}_{(f,h)}(\mathcal{X}\times \mathcal{Z})}_{\text{Language } \mathcal{X} \text{ and vision}} + \underbrace{\mathcal{L}_{(g,h)}(\mathcal{Y}\times\mathcal{Z}),}_{\text{Language } \mathcal{Y} \text{ and vision}}
\label{eq:general_model}
\end{equation}
where $\mathcal{L}$ is a metric-learning loss between text and video embeddings~\cite{miech2019end}. 
The parameters $f$, $g$, and $h$ define the embedding functions of the language $\mathcal{X}$, language $\mathcal{Y}$,  and the video domain $\mathcal{Z}$, respectively.
The idea is that \emph{sharing} the visual encoder $h$ across the two languages is crucial to align the two languages $\mathcal{X}$ and $\mathcal{Y}$.

Next, we describe the proposed approach (\Equation{eq:general_model}) in detail.
\Section{sec:architecture} explains our choice of embedding models $f$, $g$ and $h$.
\Section{sec:joint_embed} defines the loss function $\mathcal{L}$. 
Finally, in \Section{sec:muve}, we explain how our initial model can be used to improve text-based word mapping techniques.

\subsection{Multilingual Visual Embedding: Architecture}
\label{sec:architecture}	
\begin{figure}[t]
	\centering
	\includegraphics[width=\linewidth]{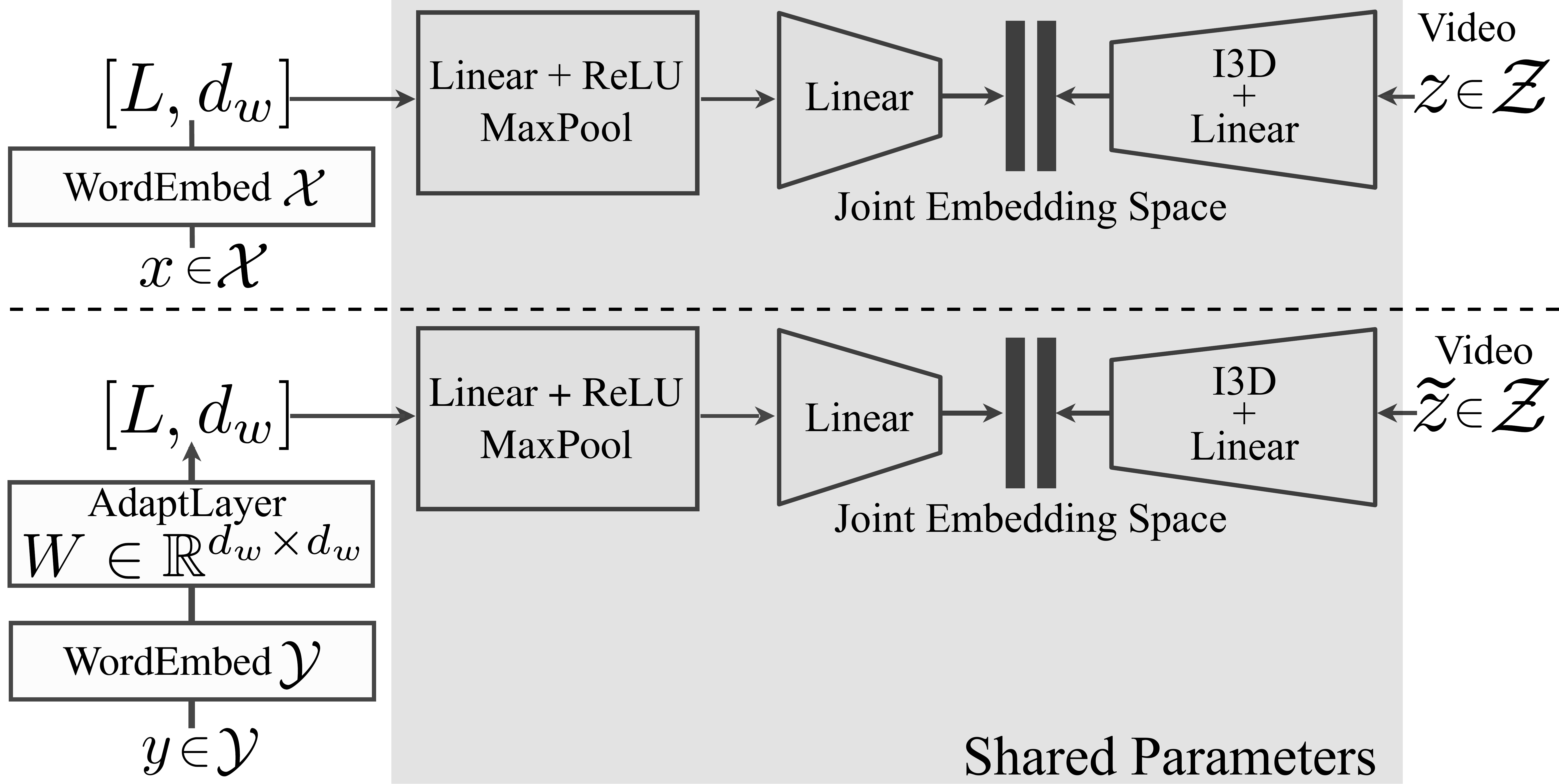}
	\caption{\label{fig:model_arch} \small Details of the three encoders: one for language $\mathcal{X}$, one for language $\mathcal{Y}$, and one for videos $\mathcal{Z}$. Coupling of the two languages is obtained by sharing parts of the model (shaded region).}
	\vspace{-0.4cm}
\end{figure}
\noindent An illustration of our architecture is given in \Figure{fig:model_arch}.

\paragraph{Input to the model.}
We represent sentences as a fixed-length sequence of integers, \ie $\mathcal{X}$ and $\mathcal{Y}$ are of the form $\{1,\ldots,K\}^L$ where $K$ and $L$ are the vocabulary size and sentence length, respectively.
On average, sentences consist of $10$ words.
Videos are in pixel space: $\mathcal{Z}=\mathbb{R}^{T\times H\times W\times 3}$, where $T$ is the number of frames in the video clips (here 32 frames at 10 FPS); $H$ and $W$ are the height and width of the video respectively, with $3$ RGB channels.

\paragraph{Text encoders.}
The text encoder $f$ in language $\mathcal{X}$, following~\cite{miech19howto100m}, consists of: (i) a word embedding layer that takes as input a sequence composed of $L$ tokens and outputs $L$ vectors of dimension $d_w$, (ii) a position-wise fully connected feed-forward layer followed by max pooling over the words to generate a single $d_i$-dimensional vector for the whole sequence, and finally (iii) a linear layer to map the intermediate representation to the joint embedding space $\mathbb{R}^d$.

For the text encoder $g$ in language $\mathcal{Y}$, we share model weights across languages~\cite{ha2016toward,johnson2017google}.
Specifically, we share the weights of the feed forward layers and the last linear layer between $f$ and $g$.
To input different languages to the shared layers, we add a linear layer, referred to as the \emph{AdaptLayer}, after the word embedding layer in language $\mathcal{Y}$.

Intuitively, the role of the \emph{AdaptLayer} is to transform the word embedding space of language $\mathcal{Y}$ such that word embeddings in language $\mathcal{Y}$ become as similar as possible to the word embeddings in language $\mathcal{X}$. Then, the rest of the network can be shared, and yet preserve the monolingual properties of the word embeddings if needed.
%
%As this layer is only linear, the model does not impose any asymmetry between languages $\mathcal{X}$ and $\mathcal{Y}$.
Our architecture is \emph{symmetric}, yet AdaptLayer appears asymmetric. However, the orthogonality constraint used in AdaptLayer enforces \emph{symmetry} of the overall model. Indeed, a symmetric case with each language equipped with AdaptLayer is equivalent to our case; this can be shown by multiplying the AdaptLayer for X and Y by the inverse of the AdaptLayer for X, ending up with a single AdaptLayer for Y.

\paragraph{Video encoder.} For the video encoder, we use the standard I3D~\cite{carreira2017quovadis} model followed by a linear layer that maps the output into the joint embedding space.

\subsection{The Base Model: Training and Inference}
\label{sec:joint_embed}

\paragraph{Training data.}
We are given a set of $n$ videos narrated in language $\mathcal{X}$:  $\{(x_i,z_i)\}_{i=1}^n$
and a set of $m$ \emph{different} videos narrated in language $\mathcal{Y}$: $\{(y_j,\tilde{z}_j)\}_{j=1}^m$.
Note that there is \emph{no} overlap in videos in the first and second set, \ie we do not have access to \emph{paired} bilingual data.

\paragraph{Training objective.}
The first term $\mathcal{L}_{(f,h)}$ in our objective function \Equation{eq:general_model} is defined as follows: 
\begin{equation}
\label{eq:loss}
\mathcal{L}_{(f,h)}\left(\{(x_i,z_i)\}_{i=1}^n\right) = \sum_i-\log \text{NCE}\left(f(x_i), h(z_i)\right),
\end{equation}
where NCE corresponds to the noise contrastive estimation~\cite{gutmann2010noise, jozefowicz2016arxiv} discriminative operator:
\begin{equation}
\label{eq:nce}
\text{NCE}(x, z) = \frac{e^{f(x)^\top h(z)}}{e^{ f(x)^\top h(z)}+\sum\limits_{(x',z')\sim\mathcal{N}}\hspace*{-4mm}e^{ f(x')^\top h(z')}},
\end{equation}
where $\mathcal{N}$ is a set of negative pairs used to enforce that video and narration that co-occur in the data are close in the space and those that do not are far.
In this work, the negatives are $x$ and $z$ paired with other $x'$ and $z'$ chosen uniformly at random from the training set $\mathcal{X}$, following~\cite{miech2019end}. In practice, each training batch includes clips from either language, and the negatives for each element in the NCE loss are the other elements from the batch in the same language.
$\mathcal{L}_{(g,h)}$ in \Equation{eq:general_model} has the same form, except with $g$ and $\{(y_j,\tilde{z}_j)\}_{j=1}^{m}$.

\paragraph{Inference.}
Because we use the same visual encoder $h$ for the two languages, we can assume that the outputs of the language encoders $f$ and $g$ are in the same space.
After training our model with the joint loss in \Equation{eq:general_model}, we can directly map the first language to the second one; for a given $x \in \mathcal{X}$, we find $y \in \mathcal{Y}$ for which the embedding $g(y)$ has the smallest cosine distance to $f(x)$.

\subsection{\vmuse{}: Improving Unsupervised Translation}
\label{sec:muve}

In this section, we explain how the Base Model can be used to improve a state-of-the-art \emph{text-based} word translation technique.

\paragraph{Text-based word translation.}
It has been shown that distributed representations of words (\eg Word2Vec~\cite{mikolov2013distributed}) share similarities across languages.
In particular, Mikolov \etal~\cite{mikolov2013exploiting} show that a word embedding matrix in a target language can be approximated by simply applying a \emph{linear} mapping on a word embedding matrix in a different source language.
To recover that linear mapping,  Mikolov \etal~\cite{mikolov2013exploiting} employ a supervised method where, given a subset of 5,000 pairs of words in the two languages, the mapping is learned by minimizing a $L_2$ distance between the word embeddings of the source language and the linearly mapped word embeddings of the target language.
Xing et al.~\cite{xing2015normalized} show that the results can be improved by adding an orthogonality constraint.
This can be done in closed form with the Procrustes algorithm (see~\cite{conneau2017word} for details).

\paragraph{The unsupervised \emph{MUSE} method.}
Conneau et al.~\cite{conneau2017word} propose the \emph{MUSE} approach that, in contrast to the method of Mikolov \etal~\cite{mikolov2013exploiting}, does not require any supervised pairs of words.
\emph{MUSE} has three main steps \textbf{(i)} finding an initial linear mapping via an adversarial approach, then \textbf{(ii)} refining the mapping with the Procrustes algorithm, and finally \textbf{(iii)} normalizing the distances using the local neighborhood.

\paragraph{\vmuse{}: aligning words through vision.}
As explained in Section~\ref{sec:architecture}, the intuition behind the linear \emph{AdaptLayer} (see \Figure{fig:model_arch}) is to map word embeddings from language $\mathcal{Y}$ to a similar vector space as word embeddings from language $\mathcal{X}$ before being fed to the shared layers.
Given this, we propose to \emph{replace} the step \textbf{(i)} (adversarial initialization) of the \emph{MUSE} algorithm by the \emph{AdaptLayer} of our Base Model, after training it on videos.
We call that method \vmuse{} for \emph{Multilingual Unsupervised Visual Embeddings}.
To further improve the performance, we follow the observation of~\cite{xing2015normalized} by adding to the objective~\eqref{eq:general_model} an orthogonal penalty $\Vert WW^\top{-}I \Vert_F^2$ on the weights $W{\in}\mathbb{R}^{d_w{\times} d_w}$ of the \emph{AdaptLayer}, where $I$ is the $d_w$-dimensional identity matrix.
In \Section{sec:vmuse_eval}, we demonstrate that \vmuse{} is more robust than its text-based counterparts in multiple aspects.

    \section{Multimodal and Multilingual Datasets}	
\label{sec:datasets}

This section explains the training and evaluation datasets used in \Section{sec:exp}. All datasets are available at \myurl{github.com/gsig/visual-grounding}.

\subsection{The \howto{} Dataset}
\label{sec:howto100m++}

Existing instructional video datasets curated from YouTube (\eg, the HowTo100M dataset) are in English. We follow the approach of~\cite{miech19howto100m} to obtain data in three new languages: French (Fr), Japanese (Ja) and Korean (Ko).
We use their list of 23,000 tasks (\eg, making a latte)
and translate them to Fr, Ja and Ko.
We obtain 31M, 30M and 34M unique clips with narration from automatic speech recognition for the Fr, Ja and Ko datasets, respectively. We use  HowTo100M~\cite{miech19howto100m} as the English (En) dataset.
To ensure that our datasets are stricly unpaired we removed any videos present in more than one of the datasets. More details are provided in the Appendix.

\subsection{Text Corpora for Training Embeddings}
\label{sec:largetext}
To compare \vmuse{} to the state-of-the-art unsupervised text-based word alignment methods, we use three text corpora:
\textbf{(i) Wiki-En/Fr}: the publicly available release of Wikipedia in English and French.
We filter the structured output to extract the sentences before processing as described in \Section{sec:impl_details},
\textbf{(ii) \howtotextall}: we use the narration extracted from the videos of \howto{} in multiple languages and
\textbf{(iii) WMT Fr-En corpus}: we use the publicly available WMT French-English corpus, that consists of En-Fr translations for various news articles.

\subsection{Evaluation benchmarks}
\label{sec:eval}

Our goal translating words from one language to another (\eg, En-Fr, En-Ko, En-Ja). We describe the datasets used to analyze the translations, also found in the Appendix.

\paragraph{The \emph{Dictionary} En-\{Fr,Ko,Ja\}.}
We use the test split of the ground-truth bilingual dictionaries used in the MUSE paper~\cite{conneau2017word} 
to compare our method to text-based word mapping methods.
Each dictionary provides the translation of $1500$ English words in another language (\eg, Fr) and list multiple translations for each English word. There are 2943 En-Fr, 1922 En-Ko, and 1799 En-Ja pairs.
As we focus on vision and to understand how different methods compare on visual versus non-visual words,
we also manually annotate the bilingual dictionary for en-fr to select words that can be visually observed (\emph{Dictionary (Visual)}). This results in 637 English words and 1430 En-Fr pairs.
Among example words in the \emph{Dictionary} dataset are: \{\emph{torpedo, giovanni, chat, catholics, herald, chuck, ...}\} whereas the \emph{Dictionary (Visual)} contains \{\emph{torpedo, chuck, pit, garrison, sprint, ...}\}.

\paragraph{\emph{Simple Words} En-\{Fr,Ko,Ja\}.}
To examine the role of word frequency, we create a list of the 1000 most common English words from the Simple English Wikipedia.
We translate this list to Fr, Ko, and Ja using the Google Translate interface. 
We manually filter these words to create a list of visual words (\emph{Simple Words (Visual)}).
Example words in the \emph{Simple Words} dataset include \{\emph{correct, touch, hit, either, regard, carry, with, three, ...}\} and \emph{Simple Words (Visual)} contains \{do, fall, police, carry, make, station, afternoon, money, club...\}

\paragraph{Human Queries En-\{Fr,Ko,Ja\}.}
In order to also qualitatively assess the performance of our proposed model in \Section{sec:qual}, we create a text dataset (\emph{Human Queries}) containing expressions similar to narrations contained in instructional videos.
We manually defined a set of 444 visual queries along with their translations in En, Fr, Ko, and Ja.
Examples include \{\emph{oil painting, make snowman, glue wood, cut tomato, play violin, open car door, paint shirt, tennis service, brew coffee, dribbling basketball, ...}\}.

    \section{Experiments}
\label{sec:exp}

In this section, we first provide our implementation details (\Section{sec:impl_details}); in \Section{sec:basemodel_eval}, we demonstrate the effectiveness of our \basemodel{} in word translation benchmarks.
In \Section{sec:vmuse_eval}, we show that the representations learned by our model can be used to improve the quality of text-based word translation methods. We also show that our method (\vmuse{}) is more robust than the text-based methods (\Section{sec:robustness}). 
Finally, in \Section{sec:qual}, we showcase various qualitative results that give further insight into our method.

\subsection{Implementation Details}
\label{sec:impl_details}

We tokenize the transcripts of the videos and lowercase. We create a vocabulary of the 65,536 most common words for each language, and map the rest to the UNK symbol.
After preprocessing, we train monolingual word embeddings using Word2Vec~\cite{mikolov2013distributed} (Skip-Gram, 300 dim, 5 words, 5 negatives).
We use these pretrained embeddings in \vmuse{}, MUSE, and VecMap models.

% DATA SAMPLING
At training, we sample a video clip (32 frames at 10 FPS) with its corresponding narration from the given datasets (\eg, \howtoen{} or the relevant \howtoshort{}-\{Fr-Ko-Ja\}). Each training batch includes clips from either language, and the negatives for each element in the NCE loss are the other elements from the batch in the same language.
% MODELS
For the video encoder, we finetune an I3D model~\cite{carreira2017quovadis} pretrained on the Kinetics-400 dataset~\cite{carreira2017quovadis}.
For the language models (\Section{sec:architecture}), the word embedding layers are pretrained on the corresponding \howtotext{} datasets to incorporate distributional semantics.
We use the Adam optimizer with an initial learning rate of $10^{-3}$ with batch size of 128 and train the model for 200k iterations on 2 Cloud TPUs.

\paragraph{Evaluation metrics.}
We report Recall@n in our experiments: given a query (\eg `Dog'), we retrieve $n$ results (\eg `Chien', `Chienne', `Chiot', \dots), and the retrieval is a success if \emph{any} of the $n$ results are listed as a correct translation in the ground-truth dictionary.
If not specified otherwise, we report \ifrone Recall@1 \else Recall@10 \fi in the paper.
We observed the same trend with \ifrone Recall@10 \else Recall@1 \fi and report it in the Appendix.

\subsection{The \basemodel{} Evaluation}
\label{sec:basemodel_eval}

We investigate whether sharing the visual encoder across languages improves the quality of word translations; to do so, we compare the results of our \basemodel{} with two baselines which we explain below.

\paragraph{Baselines.}
Our first baseline method (\emph{Random Chance}) retrieves a random hypothesis translation without using videos.
The second baseline -- \emph{Video Retrieval} -- uses videos to create a parallel corpus between the two languages. We first extract I3D features pretrained on Kinetics~\cite{carreira2017quovadis} for all video clips in \howtoen{} and \howtofr{}.
We then, for each of the English video clips (100M), find the three closest French video clips (in terms of the L2 distance). 
Finally, we take the narrations associated with these video pairs to create a parallel text corpus. Given the parallel corpus, we can find alignments between English and French words based on their co-occurrence.
More specifically, we calculate the joint probability between the English and French word pairs. 
For each English word, we can then rank the French words using this joint probability.

\begin{table}[]
    \tablestyle{6pt}{1.05}
	\begin{tabularx}{\linewidth}{p{1em}Xlccccc}
		\toprule
		\multicolumn{2}{l}{English-French} & \multicolumn{2}{c}{Dictionary} && \multicolumn{2}{c}{Simple Words} \\
		\cmidrule{3-4}
		\cmidrule{6-7}
		&& All & Visual && All & Visual \\
		\midrule
		\ifrone
		1) & Random Chance & \z0.1 & \z0.2 && \z0.1 & \z0.2 \\
		2) & Video Retrieval & \z6.3 & \z7.6 && 12.5 & 18.6  \\
		3) & \basemodel{} & \z9.1 & 15.2 && 28.0 & 45.3 \\
		4) & \vmuse{} & \textbf{28.9} & \textbf{39.5} && \textbf{58.3} & \textbf{67.5} \\
		\else
		1) & Random Chance & \z0.7 & \z1.6 && \z1.0 & \z2.2 \\
		3) & \basemodel{} & 21.3 & 33.7 && 54.0 & 74.0 \\
		4) & \vmuse{} & \textbf{45.7} & \textbf{60.7} && \textbf{79.7} & \textbf{89.5} \\
		\fi
		\bottomrule
	\end{tabularx}
	\caption{The performance of our models and the baselines as \ifrone Recall@1 \else Recall@10 \fi on En-Fr \emph{Dictionary} and \emph{Simple Words}.}
	\label{tab:vision}
	\vspace{-0.4cm}
\end{table}

\paragraph{Results.}
We report the results of our models and the baselines on the Dictionary and Simple Words benchmarks in \Table{tab:vision}. 
We observe that our \basemodel{} outperforms the baseline by a significant margin in both benchmarks. Moreover, not surprisingly, the performance of all methods is better on the \emph{Visual} portion of these benchmarks.
In \Figure{fig:noisy}, we provide two examples of the two types of failures of the \emph{Video Retrieval} model: 
In the first row, the retrieved video is correct (visually related to the query) but the narrations in English and French do not convey the same meaning. In the second row, the frame from the retrieved video is somewhat visually similar to the query (both contain food) but does not depict the same concept. This example shows how visual understanding poses a challenge for this task.

\newcommand{\tblfigtxt}[3]{
	\begin{tabular}{@{}c@{}}
		\begin{minipage}{0.48\linewidth}
			\centering
			% aspect ratio: 2.37
			\includegraphics[width=1.0\linewidth,height=6em]{#1}
		\end{minipage} \\[2.7em]
		\scriptsize{\emph{``#2''}} \\
		\scriptsize{\emph{#3}}
	\end{tabular}
}

\begin{figure}[]
	\tablestyle{0pt}{.7}
	\begin{tabularx}{\linewidth}{cXcc}
	    \toprule
		Video in \howtoen{} && Nearest Video in \howtofr{} \\
		\midrule
		\tblfigtxt{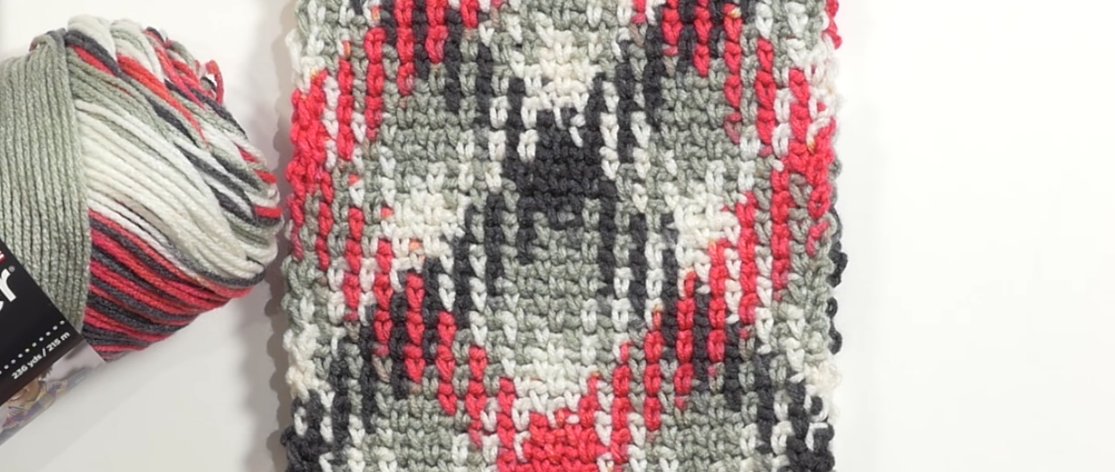}{...stich getting color sequence...}{} &&
		\tblfigtxt{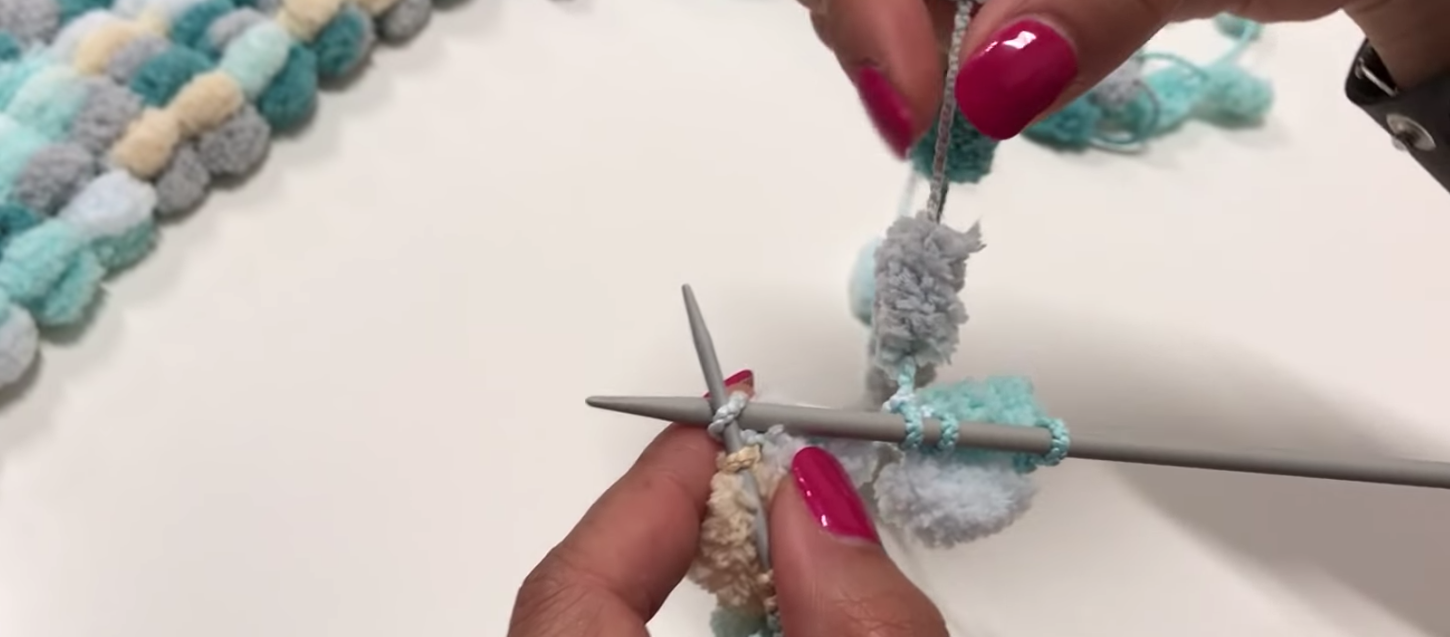}{...le pompon va se placer...}{(...the pompom will be placed...)}
		\\[1cm]
		\tblfigtxt{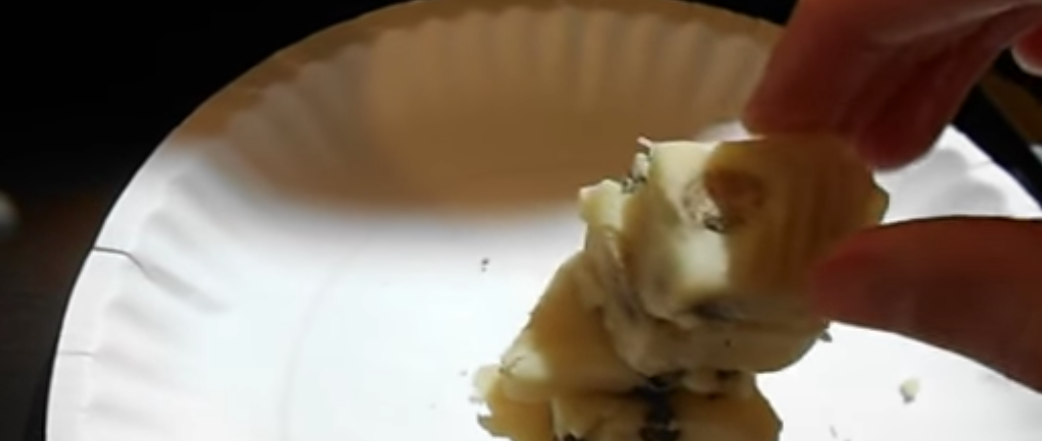}{...thank you for watching bye bye...}{} &&
		\tblfigtxt{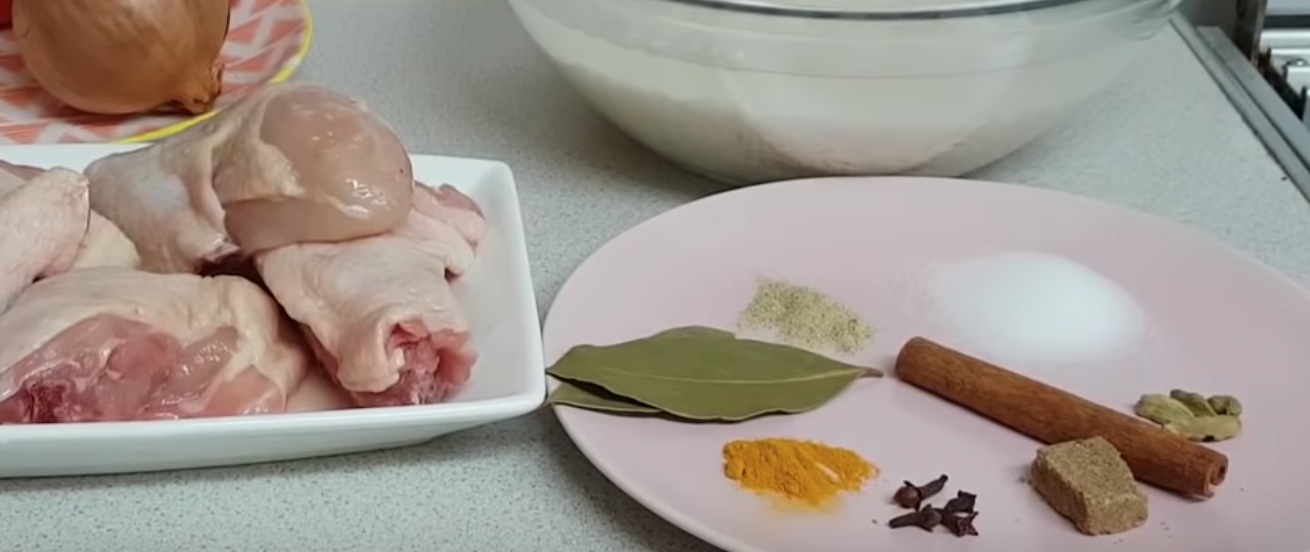}{...j'ai besoin de curcuma et de clous...}{(...I need turmeric and cloves...)} \\
		\bottomrule
	\end{tabularx}
	\caption{Examples of two types of failures of the \emph{Video Retrieval} baseline. In the first row, the videos are visually related (knitting), but no words match, making learning translation challenging. In the second, the videos are related (food), but the left caption is irrelevant to the visual content.}
	\label{fig:noisy}
	\vspace{-0.3cm}
\end{figure}

    \subsection{\vmuse{}: Improving Text-Based Alignment}
\label{sec:vmuse_eval}

We evaluate the proposed \vmuse{} approach, that is how much the representations learned by our \basemodel{} can improve the text-based word translation methods.
We first describe text-based methods that use large scale corpora for word translation. Then, we show how using representations from our model (\Section{sec:architecture}) improves over text-based approaches: three unsupervised, and one supervised methods described below. 
All methods use word embeddings trained on \emph{\howtotext{}} for their respective languages.

\noindent\emph{Iterative Procrustes} iteratively maps word embeddings of two languages using a distance-based heuristic; then it finds 
the orthogonal matrix that best maps the chosen pairs. 
We choose the best solution from 25 different initializations (either the identity matrix or random matrices).

\noindent\emph{MUSE}~\cite{conneau2017word} uses adversarial training to map the word embeddings to a space where they are indistinguishable, which provides better starting point for the \emph{Iterative Procrustes} method. The results obtained from \emph{MUSE}~\cite{conneau2017word} have been found to be sensitive to the initialization~\cite{artetxe2017acl}.

\noindent\emph{VecMap}~\cite{artetxe2017acl} is more robust to initialization and differences across languages when compared to \emph{MUSE}; it obtains better linear transformation by careful normalization, whitening, and dimensionality reduction.

\noindent\emph{Supervised} provides an upper bound on the unsupervised methods: it uses 5,000 words and their translations to find an optimal orthonormal matrix that aligns the embeddings.

\paragraph{Results.} 
In Table~\ref{tab:vanilla}, we present the word translation results between English and French, Korean, and Japanese. 
Our method, \vmuse{}, outperforms all the text-based methods. 
We observe a bigger improvement over the text-based methods for English-Korean and English-Japanese pairs. 
These results confirm previous findings that suggest text-based methods are more suited for similar languages (\eg, English and French)~\cite{sogaard2018limitations,artetxe2017acl} and shows that grounding in visual domain for word translation is especially effective in that regime.
Finally, we also observe in Table~\ref{tab:vision} a significant improvement of \vmuse{} (row 4) over our Base model alone (row 3) (+$19.8\%$ and +$30.3\%$ absolute improvement on the Dictionary and Simple Words benchmarks, respectively).
Overall, this experiment validates our intuition that the information contained in the visual domain is \emph{complementary} to the word co-occurence statistics used by the text-based methods for the task of unsupervised word translation.

\paragraph{Importance of the orthogonal constraint.}
As explained in \Section{sec:muve}, we add an orthogonal constraint to the \emph{AdaptLayer} when applying \vmuse{}.
We observe that this penalty was a \emph{key} component for \vmuse{}.
Precisely, there is a $43.0\%$ relative drop of performance for Recall@1 on the Dictionnary En-Fr (going from 28.9 in \Table{tab:vanilla} to 16.6) benchmark when removing the orthogonal constraint.
This further corroborates the findings described in~\cite{xing2015normalized}.

\begin{table}[]
    \tablestyle{4pt}{1.05}
	\begin{tabularx}{\linewidth}{p{1em}Xlcccccc}
		\toprule
		\multicolumn{2}{l}{\multirow{2}{*}{Dictionary}} & \multicolumn{2}{c}{En-Fr} && En-Ko && En-Ja \\
		\cmidrule{3-4} \cmidrule{6-6} \cmidrule{8-8}
		& & All & Visual && All && All \\
		\midrule
		\ifrone
			1) & Iterative Procrustes                &  \z0.2         &\z0.3          && \z0.3         && \z0.3 \\
			2) & MUSE~\cite{conneau2017word}  &  26.3          & 36.2          && 11.8          && 11.6 \\
			3) & VecMap~\cite{artetxe2017acl}        &  28.4          & \textbf{40.8}          && 13.0          && 13.7 \\
			4) & \vmuse{}                            &  \textbf{28.9} & 39.5 && \textbf{17.7} && \textbf{15.1} \\
			\midrule
			5) & Supervised                          &   57.9         & 60.3          && 41.8          && 41.1 \\
		\else
			1) & Iterative Procrustes                &  \z0.8         &\z0.9          && \z1.1         && \z0.6 \\
			2) & MUSE~\cite{conneau2017word}  &  42.3          & 57.8          && 23.9          && 23.5 \\
			3) & VecMap~\cite{artetxe2017acl}        &  44.1          & 60.7          && 26.8          && 27.3 \\
			4) & \vmuse{}                            &  \textbf{45.7} & \textbf{60.7} && \textbf{33.4} && \textbf{31.2} \\
			\midrule
			5) & Supervised                          &   80.1         & 84.0          && 72.1          && 68.3 \\
		\fi
		\bottomrule
	\end{tabularx}
	\caption{Performance of our and text-based methods across different language pairs. We report \ifrone Recall@1 \else Recall@10 \fi on the \emph{Dictionary} dataset. All method use word embeddings trained on \emph{\howtotext{}} for their respective languages.}
	\label{tab:vanilla}
	\vspace{-0.3cm}
\end{table}

    \subsection{Robustness of Unsupervised Word Translation}
\label{sec:robustness}
	
\Section{sec:vmuse_eval} shows that \vmuse{} is more robust to the difference between language pairs  when compared to the text-based methods (\ie performance degrades less when going from French to Japanese and Korean in \Table{tab:vanilla}). 
Here we examine two other axes of robustness: the dissimilarity of the training corpora of the two languages and the amount of training data. 
All results reported in this section are on English and French languages because text-based models perform better for this pair.

\paragraph{Model selection.}
We observe that MUSE~\cite{conneau2017word} and VecMap~\cite{artetxe2017acl} are both sensitive to initialization.
To address this, we select the optimal hyperparameters for the text-based method on the \emph{test set}: we perform an extensive search over hyperparameters and random initialisations, \eg 213 runs for the \emph{MUSE} method, and compute the performance of these runs. 
We then select the \emph{best} performing run on the test set, and hence reporting an upper bound of the true performance of these baselines.
Note that when reporting numbers for \vmuse{} we \emph{only} use the monolingual validation loss for model selection, and all numbers for \vmuse{} use the same hyperparameters.

\paragraph{Dissimilarity of the training corpora.} 
We examine how the dissimilarity of the training corpora affects the models. 
Following~\cite{eisenschlos2019multifit} we measure the dissimilarity of two corpora by comparing their word co-occurrence statistics.
Specifically, we count the co-occurrence of each pair of words in the same sentence, and normalize to get a distribution per word. Then, we align pair of words in English and French using the Google Translate interface, and compute the Jensen Shannon distance between the distributions.	

We report the results in \Table{tab:text_distribution}; all methods are evaluated on the Dictionary dataset with the Recall@10 metric.
Looking at the diagonal of the table, we observe when the corpora are \emph{similar} (\eg, Wiki-En and Wiki-Fr), all methods perform well.
However when the corpora are \emph{less similar} (off-diagonal elements), we observe that \vmuse{} significantly outperforms its text-based counterparts.
We note that methods trained on Wiki-En and WMT-Fr perform better compared to Wiki-Fr and WMT-En.
This is likely due to the combination of Wiki-Fr and WMT-En being a smaller corpora: Wiki-En is much larger than Wiki-Fr while the WMT corpora in both languages are of the same size.
In conclusion, our method by using visual grounding is more robust to the dissimilarity of the corpora in two languages.

\begin{table}[]
    \tablestyle{1pt}{1.05}
    \scriptsize
    \begin{tabularx}{\linewidth}{lXl@{\hskip .5em}cccXl@{\hskip .5em}cccXl@{\hskip .5em}ccc}
        \toprule
        && \multicolumn{4}{c}{\footnotesize{\howtofr{}}}
        && \multicolumn{4}{c}{\footnotesize{WMT-Fr}}
        && \multicolumn{4}{c}{\footnotesize{Wiki-Fr}}
        \\ \cmidrule{3-6}\cmidrule{8-11}\cmidrule{13-16}
        && {\color[HTML]{666666}$\sim$} & \cite{conneau2017word} & \cite{artetxe2017acl} & \tiny{MUVE}
        && {\color[HTML]{666666}$\sim$} & \cite{conneau2017word} & \cite{artetxe2017acl} & \tiny{MUVE}
        && {\color[HTML]{666666}$\sim$} & \cite{conneau2017word} & \cite{artetxe2017acl} & \tiny{MUVE}
        \\ \midrule
        \footnotesize{HTW-En}
            && {\color[HTML]{666666}.62}  & 45.8 & 45.4   & \textbf{47.3} 
            && {\color[HTML]{666666}.67}        & \z0.3           & \z0.7    & \textbf{35.1} 
            && {\color[HTML]{666666}.65}        & \z0.3  & \z0.1           & \textbf{41.2} 
            \\
        \footnotesize{WMT-En}     
            && {\color[HTML]{666666}.54}        & \z0.3  & \z0.2    & \textbf{26.4} 
            && {\color[HTML]{666666}.40}        & \textbf{88.0} & 87.2   & 85.0          
            && {\color[HTML]{666666}.44}        & 45.9 & \z1.3           & \textbf{54.9} 
            \\
        \footnotesize{Wiki-En}     
            && {\color[HTML]{666666}.54}           & \z0.3  & \z0.1    & \textbf{32.6} 
            && {\color[HTML]{666666}.46}      & \textbf{56.7} & 52.3   & 55.9          
            && {\color[HTML]{666666}.39} & 86.2 & \textbf{86.7} & 82.4  
            \\
        \bottomrule
    \end{tabularx}
	\caption{
	Robustness of different methods to the dissimilarity of training corpora. We report Recall@10 on \emph{English-French Dictionary} dataset for \emph{MUSE}~\cite{conneau2017word}, \emph{VecMap}~\cite{artetxe2017acl}, and \emph{MUVE}, as well as the dissimilarity ($\sim$) of the training corpora expressed with the Jensen Shannon Distance.}
	\label{tab:text_distribution}
\end{table}

\paragraph{Amount of training data.}
\label{sec:scarcity}
Unsupervised word translation is especially appealing for low-resource languages where there is no large corpora available.
We investigate to what extent \vmuse{} and the text-based methods are robust to the varying size of training data. 
More specifically, we use 100\%, 10\%, and 1\% of the target training corpora (Wiki-Fr or \howtofr) and report Recall@10. 
For \vmuse{}, when reducing \howtofr, we also reduce the amount of videos processed.
Our results are shown in \Figure{fig:scarcity}. 
\vmuse{} is more robust to conditions where the training corpora is small when compared to the text-based methods, revealing another advantage of visual grounding for the task of unsupervised word translation.

\begin{figure}
    \centering
    \includegraphics[width=1.0\linewidth]{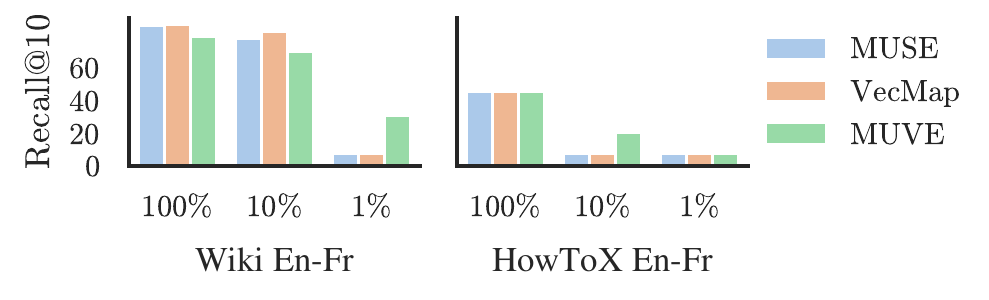}
    \vspace*{-0.7cm}
	\caption{
	Recall@10 on \emph{En-Fr Dictionary} for \emph{MUSE}, \emph{VecMap}, and \emph{\vmuse{}} varying amount of data. 
	}
	\label{fig:scarcity}
	\vspace{-0.3cm}
\end{figure}

\paragraph{Vocabulary size.}
The text-based methods rely on words' context to align the space of two languages; consequently, the size of vocabulary (and the number of words' neighbors) can play a role in their performance. 
For low-resource languages, we do not have access to a large corpus and as a result words might not have many neighbors. We explore to what extent the vocabulary size influences the performance of different methods. \Figure{fig:vocab} shows Recall@10 for different methods and vocabulary sizes. We keep the full English vocabulary and vary the size of French vocabulary. We only evaluate on words that are seen in both English and French vocabularies. 
We observe that \vmuse{} is the only method whose performance does not deteriorate when vocabulary size decreases (even when it is as small as 500).

\begin{figure}
    \centering
    \includegraphics[width=1.0\linewidth]{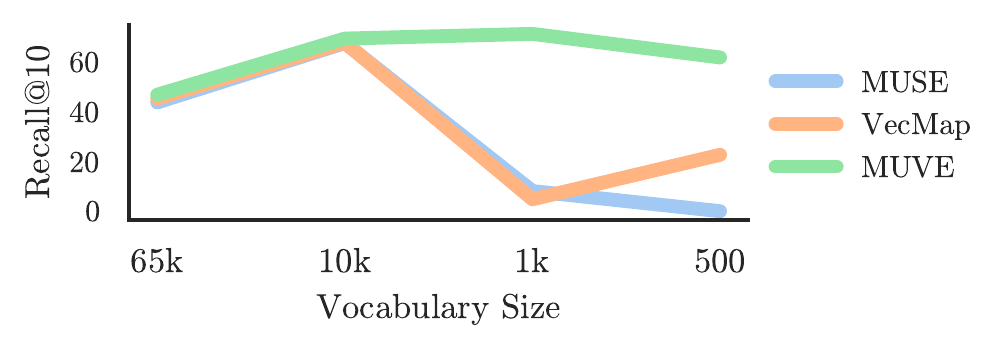}
    \vspace*{-0.7cm}
	\caption{Recall@10 on \emph{English-French Dictionary} for  \emph{MUSE}, \emph{VecMap}, and \emph{\vmuse{}} for English and French pretrained word embeddings with various vocabulary sizes in French (65k, 10k, 1k, or 500 most common French words). All methods use \howtoen{} and \howtofr{}.}
	\label{fig:vocab}
	\vspace*{0.1cm}
\end{figure}

    \subsection{Qualitative Results}
\label{sec:qual}

In \Figure{fig:reasoning}, we visualize a 2-stage inference process: \textbf{(1)} given an English query (from the \emph{Human Queries} dataset), using our \basemodel, we retrieve the video from the training set that is most similar to that query.
\textbf{(2)} Given that video, we retrieve the closest text from the French Human Queries dataset.
The model is able to retrieve relevant videos.
However, we also observe that such 2-stage approach can be problematic for translation (\eg the second row of \Figure{fig:reasoning} where both individual steps makes sense but the overall result is incorrect due to model drift).

In Table~\ref{tab:qualitative}, we  visualize the 1-stage inference process described in \Section{sec:joint_embed}.
The model is often accurate, and errors often result in semantically similar words, such as translating \emph{``a man with a dog''} as \emph{``walk dog''} and \emph{``feed dog''}.

\newcommand{\tbltxt}[2]{
	\begin{tabular}{@{}ll@{}}
		\tiny{ #1 } \\
		\ \ \ \scriptsize{(\emph{#2})}
	\end{tabular}
}
\newcommand{\tblfig}[1]{
	\begin{minipage}{0.2\linewidth}
		\centering
		\vspace{.25em}
		\includegraphics[width=1.0\linewidth,height=3em]{#1}
	\end{minipage}
}
\begin{figure}[]
	\tablestyle{1pt}{1.05}
	\scriptsize
	\begin{tabularx}{\linewidth}{lcXll}
		\toprule
		\footnotesize{English} & \footnotesize{Retrieved Video} && \multicolumn{2}{r}{\footnotesize{Top French Hypotheses Given Video}} \\
		\midrule
		Beach 
		& \tblfig{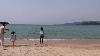}
		&& \tbltxt{Plage}{Beach}
		& \tbltxt{Courir sur la plage}{Running on the beach}
		\\
		Point at the sky 
		& \tblfig{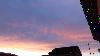}
		&& \tbltxt{Des Nuages}{Clouds}
		& \tbltxt{Le coucher du soleil}{Sunset}
		\\
		Christmas tree 
		& \tblfig{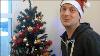}
		&& \tbltxt{Sapin de no\"el}{Christmas Tree}
		& \tbltxt{Faire un bonhomme de neige}{Make snowman}
		\\
		Cut carrot
		& \tblfig{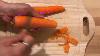}
		&& \tbltxt{Couper la carotte}{Cut carrot}
		& \tbltxt{Carotte}{Carrot}
		\\
		Add pickle
		& \tblfig{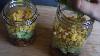}
		&& \tbltxt{Ajoutez des cornichons}{Add pickle}
		& \tbltxt{M\'elanger les legumes verts}{Mix greens}
		\\
		Add water
		& \tblfig{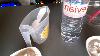}
		&& \tbltxt{Verser de l'eau}{Pour water}
		& \tbltxt{Bien m\'elanger}{Mix thoroughly}
		\\
		\bottomrule
	\end{tabularx}
	\vspace*{-0.2cm}
	\caption{
	 Left: a frame from the video that the model chose as most related to the english query. Right: top 2 french predictions conditioned on the video. The visual grounding provides a weak but usable signal for translation.}
	\label{fig:reasoning}
	\vspace{0.3cm}
\end{figure}

\newcommand{\tbltx}[2]{
	\begin{tabular}{@{}ll@{}}
		#1 \\
		\ \ \ \emph{#2}
	\end{tabular}
}
\begin{table}[]
    \tablestyle{1pt}{1.05}
    \scriptsize
	\begin{tabularx}{\linewidth}{lll}
		\toprule
		\tbltx{\footnotesize{English Text}}{} &
		\tbltx{\footnotesize{1st Model Retrieval}}{\footnotesize{(English Meaning)}} &
		\tbltx{\footnotesize{2nd Model Retrieval}}{\footnotesize{(English Meaning)}} \\
		\midrule
		\tbltx{Boy Playing}{} &
		\tbltx{Balle qui rebondit par le chat}{(Ball Bouncing by the Cat)} &
		\tbltx{Homme jouant au foot}{(Man Playing Football)}
		\\
		\tbltx{Girl Eats Ice Cream}{} &
		\tbltx{Chocolat}{(Chocolate)} &
		\tbltx{Sucrer les pancakes}{(Top Pancake Sugar)}
		\\
		\tbltx{Man Driving Red Car}{} &
		\tbltx{Homme conduit voiture rouge}{(Man Driving Red Car)} &
		\tbltx{Voiture rouge}{(Red Car)}
		\\
		\tbltx{A Man with a Dog}{} &
		\tbltx{Promener un chien}{(Walk Dog)} &
		\tbltx{Nourrir un chien}{(Feed Dog)}
		\\
		\tbltx{Air Conditioning}{} &
		\tbltx{Voler dans les airs}{(Fly Air)} &
		\tbltx{Air conditionn\'e}{(Air Conditioning)}
		\\
		\bottomrule
	\end{tabularx}
	\vspace*{-0.2cm}
	\caption{Top 2 retrieved results in French on the \emph{Human Queries} dataset given an English query.}
	\label{tab:qualitative}
	\vspace{0.2cm}
\end{table}

    \section{Conclusion}
    
    Learning multiple languages is a challenging problem that multilingual children tackle with ease. The shared visual domain can help as it allows children to relate words in different languages through the similarity of their visual experience. Inspired by this, we propose an unsupervised multimodal model for word translation that learns from instructional YouTube videos. This is beneficial over text-based methods, allowing for more robust translation when faced with diverse corpora.
    Future work needs to explore extensions to the proposed model for translating full sentences. %and move beyond automatic speech recognition to work directly with audio. 
    
    \vfill
    {\small
    \paragraph{Acknowledgements.}
    The authors would like to thank Antoine Miech for invaluable advice and their HowTo100M code, as well as Carl Doersch, Ankush Gupta, Relja Arandjelović, Viorica Pătrăucean, Ellen Clancy, and others at DeepMind for helpful discussions, support, and feedback on the project. The authors would finally like to thank Lisa Anne Hendricks and Sebastian Ruder for their feedback on the manuscript. 
    }

    \newpage
    {\small
    \bibliographystyle{ieee_fullname}
    \bibliography{cvpr2020gunnar}
    }

    \ifappendix
        \newpage
        \section{Appendix}
        
    \fi
\fi  % supplementary
\end{document}